\documentclass[10pt,twocolumn,letterpaper]{article}

\usepackage[pagenumbers]{iccv} 

\usepackage{float}
\usepackage{placeins}

\definecolor{iccvblue}{rgb}{0.21,0.49,0.74}
\usepackage[pagebackref,breaklinks,colorlinks,allcolors=iccvblue]{hyperref}
\usepackage{multicol}
\usepackage{booktabs}
\usepackage{multirow}
\usepackage{array}
\usepackage{adjustbox}
\usepackage{tabularx}

\newcommand{\methodname}{GAS-NeRF\xspace}

\title{\methodname: Geometry-Aware Stylization of Dynamic Radiance Fields}

\author{Nhat Phuong Anh Vu$^{* 1}$~\quad Abhishek Saroha$^{* 1,2}$~\quad Or Litany$^{3,4}$~\quad Daniel Cremers$^{1,2}$
\\[0.2ex]
\small{$^{1}$Technical University of Munich} \quad 
\small{$^{2}$Munich Center for Machine Learning} \quad 
\small{$^{3}$Technion} \quad 
\small{$^{4}$Nvidia}\\
\small{$*$ denotes equal contribution}\\
}

\begin{document}
\twocolumn[{
\maketitle
\begin{center}
    \captionsetup{type=figure}
    \includegraphics[width=0.91\textwidth]{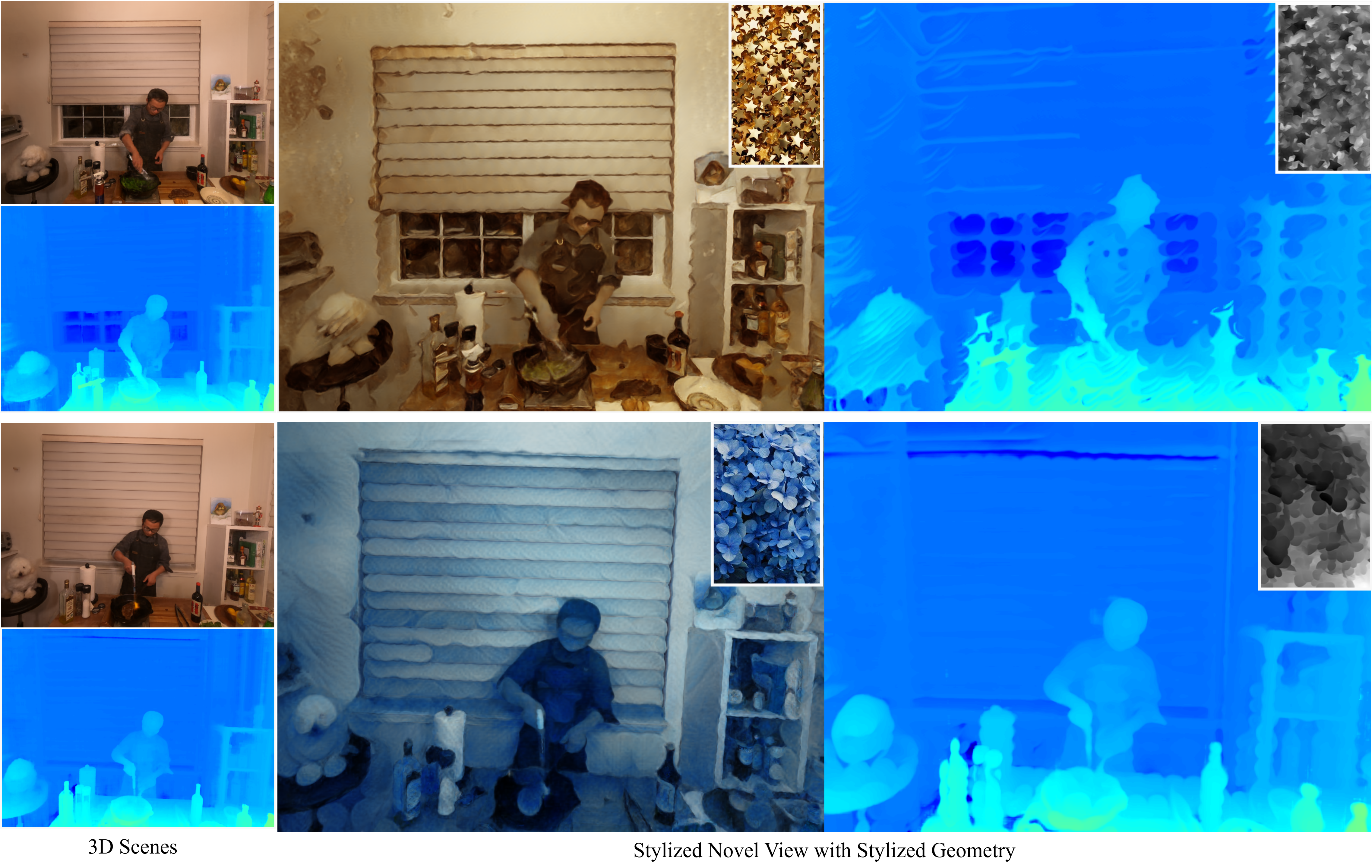}
    \caption{We present a method to perform stylization of a given dynamic scene based on an input style image. We optimize the geometry in conjunction with the appearance to make the stylization more physically plausible, such that the appearance and the underlying geometry complement each other. To the best of our knowledge, we are the first method to perform geometry transfer for dynamic radiance fields.}
    \label{fig:teaser}
\end{center}
}]

\begin{abstract}
Current 3D stylization techniques primarily focus on static scenes, while our world is inherently dynamic, filled with moving objects and changing environments. Existing style transfer methods primarily target appearance -- such as color and texture transformation -- but often neglect the geometric characteristics of the style image, which are crucial for achieving a complete and coherent stylization effect. To overcome these shortcomings, we propose \methodname, a novel approach for joint appearance and geometry stylization in dynamic Radiance Fields. Our method leverages depth maps to extract and transfer geometric details into the radiance field, followed by appearance transfer. Experimental results on synthetic and real-world datasets demonstrate that our approach significantly enhances the stylization quality while maintaining temporal coherence in dynamic scenes.
\end{abstract}
\section{Introduction}


\begin{figure*}[h]
    \centering
    \includegraphics[width=\textwidth]{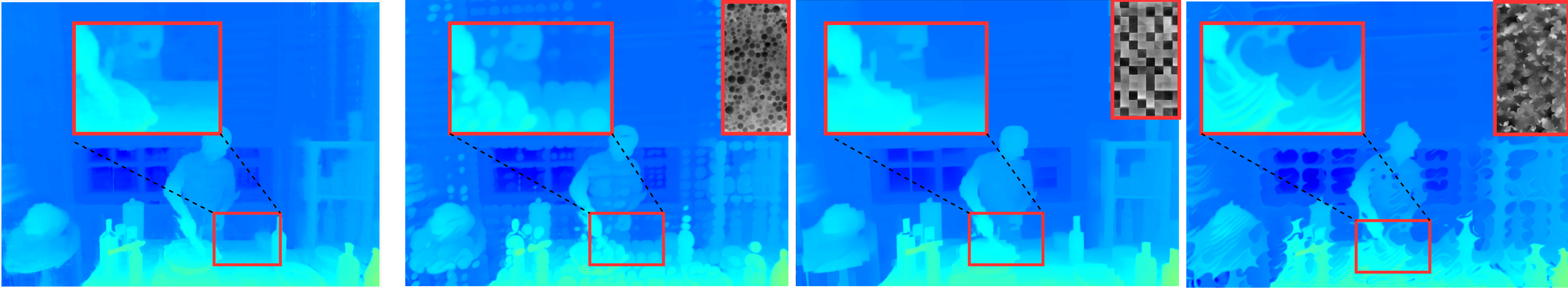}
    \caption{\textbf{Comparison of stylized geometry before and after stylization}. We show the effect of various different kinds of style images on the geometry of the underlying scene after our method. It can be seen that the underlying geometry, on the left, is adapted to match that of the depth of the input style images, which are geometrically different from each other, to provide a more comprehensive overview of the ability of our method. } 

    \label{fig:depth_stylization}

\end{figure*}

Style, or the art of imagining things in certain way, forms the strong backbone of human creativity. Whether in painting, photography, films, or user interfaces, style enhances aesthetic appeal and provides deeper meaning to visual and sensory experiences. Inspired by this very idea, Style transfer is a technique in computer vision that applies the artistic characteristics-such as color, texture, and brush strokes-of one image (the style image) to another image (the content image). This method has a wide range of applications across digital art, entertainment, augmented reality, virtual reality, computer gaming, to name a few.

With the advent of deep learning and neural networks, the pioneering work of Gatys \etal ~\cite{gatys2016image} opened up a realm of possibilities, namely Neural Style Transfer, making it possible to imagine how any given image would look like, from the creative eyes of famous painters. This was, however, not the first time style transfer was explored as a research problem. Earlier works on texture transfer \cite{efros1999texture,wei2000fast,kwatra2003graphcut, efros2023image}
could also be perceived as a parallel line of research on style transfer. Following ~\cite{gatys2016image}, many more learning-based methods, such as ~\cite{johnson2016perceptual,huang2017adain} offered a more flexible and robust approach to neural style transfer. These methods were trained on a large corpus of training images, and therefore, could in theory, generate the stylized images in one forward pass, as opposed to the earlier work of ~\cite{gatys2016image} that relied on optimizing the set of input style and content images.

Recently, more methods also focused on improving the representation of the 3D scenes. Radiance Fields(RF), in particular, Neural Radiance Fields(NeRFs)~\cite{mildenhallnerf}, was at the helm of this development.  As a result, extending style transfer techniques to 3D scenes emerged as a natural area of research. The works of ~\cite{huang2022stylizednerf,nguyenphuoc2022snerf,arf,zhang2023ref} built on top of a neural radiance field, to obtain a stylized 3D scene representation conditioned on a style image. Similar to image-based style transfer, most works  ~\cite{nguyenphuoc2022snerf,arf,zhang2023ref} focused on optimizing a radiance field given each style image, while the works of ~\cite{huang2022stylizednerf} trained a network to predict the colors of each queried 3D point coupled with a style image to obtain the stylized radiance field. While optimization-based methods achieved better stylization quality, StylizedNerf, and the following works of zero-shot stylization ~\cite{liu2023stylerf} made them more accessible and easier to adapt for all practical use-cases. A similar trend was seen with the arrival of 3D Gaussian Splatting~\cite{kerbl3Dgaussians}, which offered real-time rendering capabilities at test-time. The works of ~\cite{liu2024stylegaussian,saroha2024gaussian,mei2025regs,ggstyle} performed scene stylization by using a Gaussian Splatting backbone. The main challenge of these static 3D scene stylization methods is enable artistic consistency in the scene, which means that a point seen from any viewing direction, should appear the same, or similar to say the least~\cite{ibrahimli2024muviecast}.

The common aim of all these methodologies is to stylize the appearance of the scene, and most of them make use of a pre-trained radiance field. Therefore, they leave the geometry unchanged. This is not an ideal scenario, because if only the appearance is changed, then it does not fit in with the actual geometry. This is undesirable for practical applications, such as game development, wherein a character needs to navigate their way through a scene or map, and the map geometry is different than its appearance. For instance, if the style image is consisting of more square-like patterns, it will not fit onto the scene consisting of balls or other round objects, which have a circular geometry. Therefore, the task of stylization should be coupled with modifying the geometry as well. Recently, the work of ~\cite{3dgeo} proposed to alter the geometry of the underlying static  scene by optimizing it using the depth map of the style image.

However, the world we live in comprises of moving objects, thereby making it dynamic in nature. Building upon the works of representing dynamic scenes~\cite{cao2023hexplane,kplanes_2023}, the works of ~\cite{li2024sdyrf,xu2024styledyrf,zdyss,liang20244dstylegaussian} extended scene stylization to dynamic scenes, performing dynamic scene stylization. The fundamental challenge in such a task is to maintain consistency across the spatial and temporal domain simultaneously. Current approaches also show limitations when dealing with moving objects, often displaying a persistent flickering effect and temporal discontinuities. Similar to that of static scenes, all of the existing works also just focus on appearance, thereby keeping the original geometry unchanged. Our approach, however, aims to address both geometry transfer and appearance stylization simultaneously. Our problem setup is aligned to that of ARF~\cite{arf}, Geo-SRF~\cite{3dgeo}, or S-DyRF~\cite{li2024sdyrf} where we focus on the optimization of the underlying scene parameters based on the conditioning style input. Similar to ~\cite{3dgeo}, we leverage the use of depth maps of the given style image to transfer the geometry to the underlying scene. In addition to depth maps, we use a combination of various loss terms to improve the stylization quality. In summary, our the contributions of our work can be summarised as follows:
\begin{itemize}
    \item Geometry-aware style transfer for dynamic 3D scenes: We incorporate geometry transfer into dynamic radiance fields, ensuring that both color and structural details from the style image are preserved.
    \item We make use of a combination of loss terms that mitigate the effects of artifacts that arise after the style transfer to the appearance and geometry of the underlying scene. 
    \item To the best of our knowledge, we are the first work to transfer the style geometry in a dynamic radiance field.
\end{itemize}

\section{Related works}

\subsection{Scene Representation using Radiance Fields}
In recent years, representing a 3D scene has been mostly focused on methods based on radiance fields. Neural Radiance Fields(NeRF) ~\cite{mildenhallnerf} was at the forefront of this revolution by encoding the scene details in the weights of a multi-layer feed forward network. Despite being a simple yet effective formulation, vanilla NeRF suffered from many drawbacks, which was tackled in several follow up works. While the works of ~\cite{reiser2021kilonerf,wang2023f2nerffastneuralradiance,yu2021plenoctreesrealtimerenderingneural,garbin2021fastnerf,muller2022instant,hedman2021bakingneuralradiancefields} focused on accelerating the training and inference speed of vanilla NeRFs, other works such as \cite{chou2024gsnerfgeneralizablesemanticneural,yu2021pixelnerfneuralradiancefields} tackled the problem of a generalized radiance field framework. Furthermore, since NeRF~\cite{mildenhallnerf} relied on dense input views covering a wide spectrum of the input scene, \cite{wang2023sparsenerf,niemeyer2021regnerf,chibane2021stereo} attempted to reconstruct a given scene with sparse inputs. Since the seminal work on NeRFs was initially designed for static scenes, \cite{pumarola2020dnerf,Gao2021DynNeRF,park2021hypernerf,park2021nerfies,cao2023hexplane} extended them to include dynamic scenes. The formulation of temporally varying scene in the form of Hexplanes~\cite{cao2023hexplane} also form the backbone of our work and we refer the reader to ~\cite{cao2023hexplane} for a more in-depth explanation. 

On the contrary, 3D Gaussian Splatting~\cite{kerbl3Dgaussians} explored an explicit, rasterization-based approach for novel-view synthesis. Similar to NeRF, it was originally proposed for static scenes, and later extended for dynamic scenes \cite{luiten2024dynamic,yang2024deformable,wu20244dgs}. Most works in this line try to model a time-varying deformable field to account for the displacement of 3D Gaussians, and hence, generally suffer from addition or removal of objects in the scene.

\subsection{Style Transfer}
The task of style transfer included transferring the style, comprised of the colors and effects such as brush strokes, to a given image. The seminal work of ~\cite{gatys2016image} achieved state-of-the-art results upon its inception. The core idea comprised of using a large neural network, such as the VGG~\cite{simonyan2014vgg}, to generate an image, that had the features of both, the style image and the given image, known as the content image. The process required optimizing the generated image for each content and style image separately, thereby limiting the practical use cases. To mitigate the effects of this costly optimization approach, methods such as \cite{johnson2016perceptual,huang2017adain} trained a feed-forward network, thereby increasing the generalization abilities for this task at hand. Recently, more sophisticated methods such as ~\cite{liu2021adaattnrevisitattentionmechanism,deng2022stytr2imagestyletransfer} made use of attention and transformer mechanism respectively to improve the stylization quality of neural networks. 

Consequently, the task of video stylization also gained popularity. It involved an additional challenge of maintaining temporal consistency, thereby reducing the effect of flickering in such stylized videos. Earlier works of \cite{chen2017coherent,huang2017real,ruder2018artisticStyleTransferVideos} relied on optical flow~\cite{dosovitskiy2015flownet} networks, while the works of ~\cite{jamrivska2019stylizingVideoByExample,texlerfspbt} made use style propagation to the video by having one or few reference stylized keyframes.

\begin{figure*}[h]
    \centering
    \includegraphics[width=\textwidth]{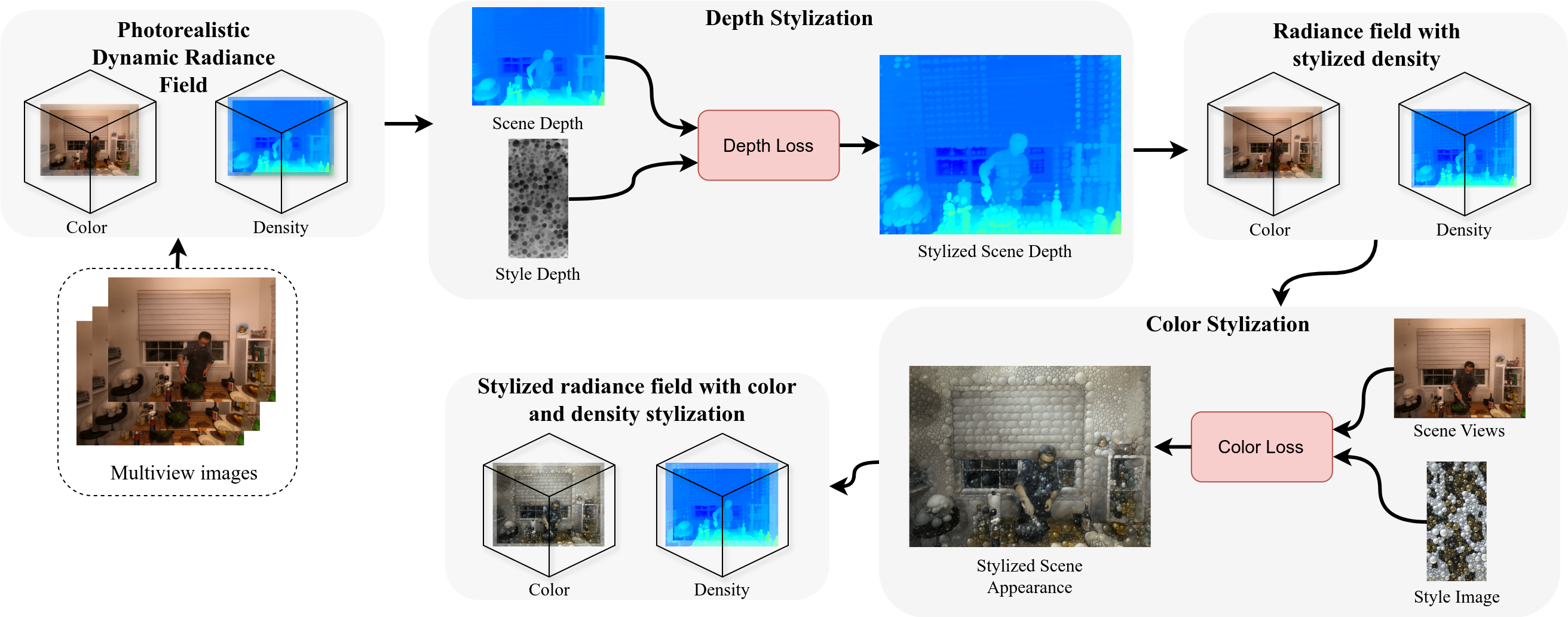}
  \caption{\textbf{Method Overview.} We start by training a photorealistic radiance field using the given multi-view input video using Hexplanes ~\cite{cao2023hexplane}. Given a style image $S_{rgb}$ and its corresponding depth map $S_{depth}$, we first freeze the appearance branch and modify the geometry(density) of the scene using $S_{depth}$, following which, we freeze the density and modify the appearance of the scene to obtain the stylized radiance field representing the scene.}
    \label{fig:overview}

\end{figure*}

\subsection{Scene Stylization}

\paragraph{Static Scene Stylization} 3D Scene Stylization refers to applying the technique of style transfer, not to an image but an entire scene. It's an inherently challenging task due to the prerequisite of maintaining spatial consistency while applying the style features to the scene. Over the years, numerous works have tried to solve this task by using different scene representations. Initial methods, such as ~\cite{huang2021lsnv,mu20223d} and ~\cite{hollein2022stylemesh} represented their scenes using point clouds and meshes respectively. Radiance fields, since their popularity, have become the preferred choice for scene stylization. SNeRF~\cite{nguyenphuoc2022snerf} trains a radiance field by optimizing the pre-trained photo-realistic radiance field with stylized training views, while preserving the underlying structure of the original scene. ARF ~\cite{arf} instead, uses a novel Nearest Neighbor Feature Matching(NNFM) loss function to transfer the style features to a pre-trained radiance field, while Ref-NPR~\cite{zhang2023ref} uses ray registration and a stylized training view to propagate the style information through the entire scene. While such approaches offer great stylization quality, they are costly to compute, since they run an optimization problem given every the scene and the input style image, thereby limiting their practical use cases. 

To mitigate the effects of this, a separate line of work focused on stylizing scenes using neural networks and a zero-shot approach by leveraging pre-trained networks. ~\cite{chen2022upstnerf,chiang2022stylizing} and ~\cite{huang2022stylizednerf}are examples of this line of research, where the methods are built on top of using a hypernetwork and training a feed-forward network to predict stylized colors based on the input style image respectively. StyleRF~\cite{liu2023stylerf} is a zero-shot method that relies on learning embedded features along with the scene, which are in turn used to transfer the style and later decoded to get the stylized scene. 

As of late, with the growing influence of 3D Gaussian Splatting, the works of ~\cite{liu2024stylegaussian,saroha2024gaussian,ggstyle,mei2025regs} explored stylizing the scene using the 3DGS framework, anticipating a better rendering speed due to the rasterization-based backbone. 

All of the above mentioned methods focus on stylizing the appearance of the scene. However, the seminal work of ~\cite{3dgeo} introduced the task of stylizing the geometry in conjunction with the appearance. This is done by modeling a deformation field to account for the changes in geometry. 

\paragraph{Dynamic Scene Stylization} Since the field of scene stylization is relatively new and unexplored, most methods only focus on static scenes. However, a few approaches such as ~\cite{li2024sdyrf,liang20244dstylegaussian,xu2024styledyrf,zdyss} explore the task of dynamic scene stylization. While S-DyRF is an optimization based approach, similar in spirit to Ref-NPR~\cite{zhang2023ref}, ~\cite{liang20244dstylegaussian,xu2024styledyrf,zdyss} focus their efforts more on the zero-shot stylization. However, none of these methods focus on modifying the underlying geometry and only focus on modifying the appearance. 

Based on our knowledge, our work is the first to focus on simultaneously stylizing both the appearance and geometry of a dynamic scene, offering a more realistic and coherent stylization approach.

\section{Methology}

On the contrary to previous scene stylization approaches that just focus on modifying the colors of the radiance field, we also want to modify the geometry, specifically for a dynamic scene. To this end, we adopt a two-step process. In the first step, we transfer the geometry from the style image, $S_{rgb}$, to our pre-trained radiance field(RF), following which we perform color transfer. We show an overview of our approach in ~\Cref{fig:overview}. 

\subsection{Dynamic Radiance Fields}
A given dynamic scene can be represented by a time-varying radiance field. The input to the method is simply videos from a multi-view setup, such that we obtain a dynamic scene with the ability to generate novel views in space and time. For this, we build up on top of Hexplanes~\cite{cao2023hexplane}. Hexplanes make use of a grid-like encoderin conjunction with MLP, conditioned on a camera view $\mathcal{C}$ and the querying time $t$ to obtain the view $I_{rgb}$ at that particular time instant. Additionally, we can also query the accompanying depth map $I_{depth}$ for that time and camera position.  Mathematically, it can be summarised as follows:

\begin{equation}
    I_{rgb}, I_{depth} = Hexplane(\mathcal{C}, t)
\end{equation}

We pretrain the radiance field using photorealistic images, before we begin transferring the geometry and style from the conditioned style image. For a more comprehensive understanding, we refer the reader to ~\cite{cao2023hexplane}. 

\subsection{Dynamic Geometry Transfer}
The obtained RF is photorealistic, \ie without any stylization effects, similar to the prior works of ~\cite{arf,zhang2023ref}. Once we have the photorealistic RF, we begin to transfer the stylized geometry onto the scene. The intuition behind performing the geometry transfer before appearance transfer instead of doing them sinultaneously in one sweep is that, once we have modified the underlying geometry according to the given style image, the appearance transfer would act as a step to "fit" the modified color values to this updated geometry, thereby leading to more coherent results. 

One crucial component of our loss function is the Nearest Neighbor Feature Matching, $NNFM$, as described in ~\cite{arf}. Originally used to distill the color-based style of the style image onto the scene, ~\cite{arf} proposed to use the $NNFM$ loss function, while Geo-SRF~\cite{3dgeo} explored the idea of using it for geometry transfer for static scenes. Given two feature maps $\mathcal{F}_{1}$ and $\mathcal{F}_{2}$, such as those obtained by passing the input images, $I_1$ and $I_2$, through a pretrained VGG~\cite{simonyan2014vgg}, $NNFM$ is mathematically defined as:

\begin{equation}
\begin{split}
    NNFM(\mathcal{F}_{1}, \mathcal{F}_{2}) = \\ \frac{1}{N} \sum_{i,j} \min_{i', j'}
    D(\mathcal{F}_{1}(i,j), \mathcal{F}_{2}(i',j'))
\end{split}
\label{eq:nnfm_loss}
\end{equation}
where $D(, )$ is the cosine distance between the two feature vectors.

For the geometry optimization, we minimize the loss function $\mathcal{L}_{depth}$ defined as follows:

\begin{equation}
\begin{split}
    \mathcal{L}_{depth} = \lambda_{nnfm} NNFM(\mathcal{F}^{Im}_{depth}, \mathcal{F}^{Style}_{depth}) \\ + \lambda_{tv} TV(I_{depth}) + \\ \lambda_{L}\mathcal{L}_{content}(I_{depth},I^{unstylized}_{depth})
\end{split}
\label{eq:depth_loss}
\end{equation}
where $NNFM()$ is the Nearest Neighbor Feature Matching, as described in ~\cite{arf} and ~\Cref{eq:nnfm_loss}, $\mathcal{F}^{Im}_{depth}, \mathcal{F}^{Style}_{depth}$ are the feature vectors obtained by passing the rendered depth map $I_{depth}$ and the depth map of the style image, $S_{depth}$, through a pretrained VGG, $I^{unstylized}_{depth}$ is the unstylized depth map from the pretrained RF obtained after the photorealistic hexplane training, $TV()$ is the Total Variation Loss, $\mathcal{L}_{content}$ is the L2 Loss, and $\lambda_{nnfm}, \lambda_{L}, \lambda_{tv}$ are the respective weight terms for each of the components of the loss function. 

We add the $NNFM$ loss so that we can effectively transfer the geometrical features of the input style image $S_{rgb}$ to our dynamic scene. The motivation to add the $\mathcal{L}_{content}$ and $\mathcal{L}_{TV}$ is to preserve the details of the original depth information, while maintaining spatial continuity respectively. The effects of both of these loss terms have been discussed in more detail in ~\Cref{sec:ablation:tv_content_losses}.


After completing the geometric style transfer process, we obtain a dynamic radiance field with stylized density.

\subsection{Gradient Scaling}
\label{sec:gradient_scaling_method}
During the process of geometry transfer to the dynamic scene, we observe visible floating artifacts being introduced. Two of the most prominent artifacts in NeRF reconstruction are floaters and excessive and unwanted reconstructed very close to the camera, also dubbed as background collapse \cite{nofloaters}. As a result, this incorrectly reconstructed geometry appears as floating artifacts when viewed from other viewpoints. The work of \cite{nofloaters} hypothesize that background collapse is primarily caused by an imbalance in gradient distribution: near-camera volumes receive a disproportionate amount of gradients due to denser sampling. This excessive gradient accumulation leads to a rapid build-up of density, resulting in floating artifacts. Given a point $p$, let  $\nabla p$ be its gradient, and let  $\delta_i p $ denote the distance between the point $p$ and the camera ray origin. We apply the following gradient scaling:

\begin{equation}
s \nabla p = \min(1, \delta_i p)
\end{equation}

In other words, we replace $ \nabla p $ with $\nabla p \times s \nabla p$. The goal is to scale down the gradients from samples near the camera, reducing their impact, while leaving the gradients from distant points unchanged. In the original work of \cite{nofloaters}, the authors use $ \min(1, (\delta_i p)^2)$ which ensures that near-camera and far-camera volumes receive the same total gradient contribution. Empirically, this approach suppresses gradients too aggressively in near-camera volumes. As a result, fine details in depth stylization become less visible. To counter this, we make use a linear gradient scaling $\min(1, \delta_i p)$ instead of a quadratic one. Our approach still reduces the excessive gradient accumulation near the camera, mitigating artifacts, but without completely equalizing the gradient distribution between near and far volumes. This allows near-camera regions to retain more gradient information, ensuring that depth stylization details remain more distinct and visible. The effect of having this gradient scaling has been discussed further in ~\Cref{sec:ablation:gradient_scaling}.

\subsection{Color Transfer}
After completing the geometry transfer, we use the radiance field with the already stylized density to perform color transfer. During this process, we freeze the density and adjust only the color.

The RGB style image, denoted as $S_{rgb}$ , serves as the style guide. Similar to the color transfer in ~\cite{arf}, we render the RGB image $I_{rgb}$ and minimize the NNFM loss between $S_{rgb}$ and $I_{rgb}$. 
Therefore, we can define the color loss, $\mathcal{L}_{color}$ as follows:
\begin{equation}
    \mathcal{L}_{style} = NNFM(I_{rgb}, S_{rgb})
    \label{eq:color_loss}
\end{equation}
where $NNFM(, )$ is similar to the one defined in \Cref{eq:nnfm_loss}. This loss function has been shown to be extremely effective and deliver state-of-the-art color transfer results while being easy to incorporate and without any significant additional overheads~\cite{arf}.
This results in a dynamic radiance field where both color and density have been stylized.

\begin{table}[htbp]
  \centering
  \begin{tabular}{p{2.2cm} cc cc}
    \toprule
    \multirow{2}{*}{Method} & \multicolumn{2}{c}{NV3D~\cite{li2022neural}} & \multicolumn{2}{c}{D-Nerf~\cite{pumarola2020dnerf}} \\
    \cmidrule(lr){2-3} \cmidrule(lr){4-5}
    & RGB $\downarrow$ & Depth$\downarrow$ & RGB$\downarrow$ & Depth$\downarrow$ \\
    \midrule
    ARF*~\cite{arf} & 6.487 & 8.809 & \textbf{8.670} & 10.149 \\
    Ref-NPR*\cite{zhang2023ref} & 6.668 & 8.809 & 8.700 & 10.149 \\
    S-Dyrf~\cite{li2024sdyrf} & \textbf{6.254} & 8.809 & 8.779 & 10.149 \\
    \midrule
    \methodname (Ours) & \underline{6.481} & \textbf{8.707} & \underline{8.687} & \textbf{10.084} \\
    \bottomrule
  \end{tabular}
  \caption{\textbf{Quantitative comparisons} Here, we show the LPIPS~\cite{zhang2018unreasonable} scores for all the baselines against our proposed method. \textbf{Lower scores indicate better similarity}. The best performance is in \textbf{bold}, and the second best is \underline{underlined}. The scores are averaged over all the scenes in each dataset. A thorough scene-wise breakdown is provided in the supplementary material. Based on the metric quantitatively, we are able to incorporate the stylized geometry from the style to our dynamic scene much better than the baselines. We can see that since all the baselines use a pretrained radiance field and do not modify the underlying geometry, they have the same LPIPS score for the depth maps. The scores are multiplied by a factor of 10 for increased readability. }
  \label{tab:lpips_avg}
\end{table}

\section{Experiments}

\subsection{Datasets and Baselines}
\label{sec:datasetsbaselines}
We used two distinct datasets for our experimentation setup: the real-world  Plenoptic Video dataset\cite{li2022neural} and the synthetic D-NeRF dataset\cite{pumarola2020dnerf}. We compare our method against three different baselines. The first is ARF\cite{arf}, a style transfer method for 3D scenes. The second baseline is Ref-NPR\cite{zhang2023ref}, a reference-based 3D stylization method. Since both of these methods were designed for static scenes, we reimplemented them to support dynamic scenes, and therefore, refer to them as ARF* and Ref-NPR* respectively. Finally, we include S-DyRF\cite{li2024sdyrf}, a reference-based stylization method specifically designed for dynamic scenes. We do not compare with Geo-SRF~\cite{3dgeo} since their code was not public at the time of writing. 

\subsection{Implementation Details}
We use Hexplane\cite{cao2023hexplane}  as the underlying NeRF representation for our dynamic radiance field, and adopt the training scheme followed in the original Hexplane paper. For feature extraction component of NNFM, we utilize the conv2 of VGG16\cite{simonyan2014vgg}. For depth stylization, we train for 1000 iterations. For color stylization, we train for 600 iterations, following the Pytorch implementation of ARF \cite{arf}. To extract the depth maps of the style image, $S_{depth}$, we use the pretrained ZoeDepth~\cite{zoedepth} network.

For the NV3D dataset, we set the depth stylization loss weights \Cref{eq:depth_loss} as $\lambda_{nnfm} = 1$, $\lambda_{tv} = 5$, and $\lambda_L = 1$. \\
For the DNeRF dataset, the weights are $\lambda_{nnfm} = 1$, $\lambda_{tv} = 5$, and $\lambda_L = 10$.

\begin{figure*}[h]
\centering
        \captionsetup{type=figure}
    \includegraphics[width=0.91\textwidth]{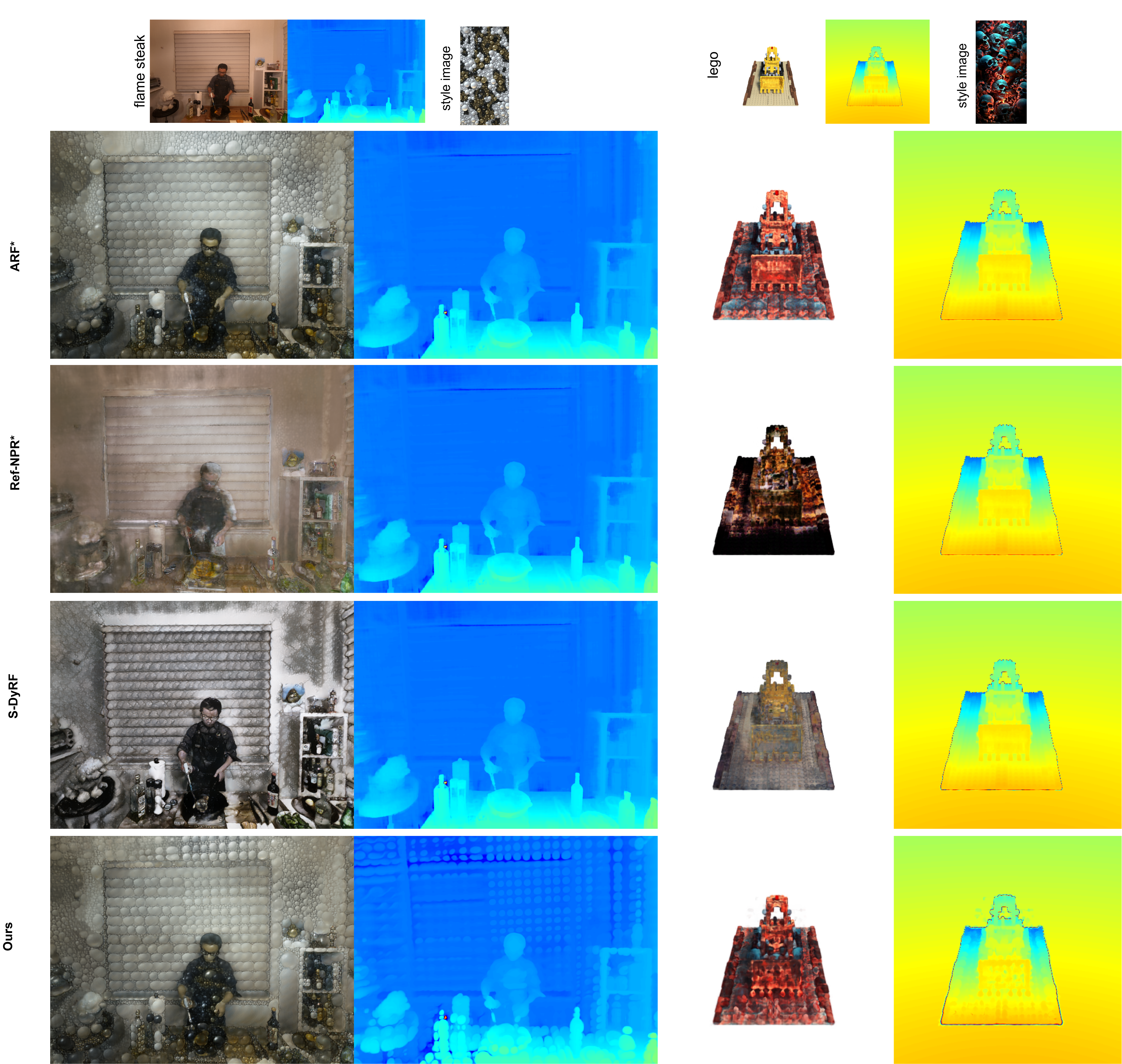}
    \caption{\textbf{Qualitative comparisons} We provide the qualitative comparisons of our method against the baselines described in ~\Cref{sec:datasetsbaselines}, namely ARF*~\cite{arf}, Ref-NPR*~\cite{zhang2023ref},S-DyRF~\cite{li2024sdyrf}. In this particular example of the Nv3D dataset, we can see that our geometry consists of circular blobs, which are expected due to the style image being made of a large number of pearls. In the case of ARF*, even though the walls and the background appear circular, the geometry of the background regions is flat and rectangular, thereby giving rise to a mismatch. The color of ours is similar to ARF* since the color transfer backbone of both the methods is similar. Similarly, in the lego scene of the D-Nerf dataset, we can see that our method modifies the underlying geometry to resemble that of the skull pattern. Ref-NPR* is able to recreate the colors in the lego scene, while it fails to maintain details in the Nv3D case. S-DyRF in particular, suffers from temporal consistency, i.e. has visible flickering effects while navigating the scene along the temporal domain. We provide accompanying videos in the supplementary to give a more comprehensive overview and comparison for the same.}
    \label{fig:qualitative}
\end{figure*}

\section{Results}

\paragraph{Qualitative comparisons}
We present the visual comparisons in \Cref{fig:qualitative}  with ARF* and Ref-NPR*, and S-DyRF~\cite{li2024sdyrf}. In our qualitative comparisons, we observe that our method significantly improves the depth map by transferring the geometric details of the style image, whereas other methods primarily focus on color transfer while neglecting depth information. This allows our approach to generate stylized outputs that maintain a stronger sense of geometric stylization. Regarding color transfer, since our method adopts a similar approach to ARF\cite{arf}, the colorization in our results is expected to be quite similar to ARF’s outputs. For S-DyRF, it can be observed that they suffer from flickering effects, especially in the temporal domain. In the supplementary materials, we provide videos to support our findings. 

\paragraph{Quantitative comparisons}
For a quantitative evaluation of all the methods, we follow the benchmark of Ref-NPR~\cite{zhang2023ref}. We compare the LPIPS \cite{zhang2018unreasonable} scores in ~\Cref{tab:lpips_avg}, which measure the perceptual difference between two images—a lower score indicates greater similarity. Specifically, we evaluate the LPIPS scores of the following setup:
\begin{itemize}
    \item RGB: The RGB LPIPS is computed between the rendered RGB Image at a particular view and timestamp, $I_{rgb}$ and the style image $S_{rgb}$.
    \item Depth: We also compute the LPIPS for the depth maps $I_{depth}$ and $S_{depth}$, which are the rendered depth map for a particular view given a time, and the depth map of the style image $S_{rgb}$. This is done to compute the similarity and effectiveness of our dynamic geometry approach.
\end{itemize}

Our method achieves state-of-the-art LPIPS score for RGB images as well as depth maps across all evaluations. The better depth score indicates that, due to geometry transfer, our method produces results that are more adhering to that of the style image. The final score was computed by generating 120 novel views of the stylized dynamic scene, varying over the spatio-temporal domain, and averaging their frame-wise values.


\paragraph{User Study}
Since there is no true metric to evaluate such methods that include stylization and geometry transfer, a standard practice for comparing and evaluating such methods is to perform a user study. For our user study, we asked the users to pick the best performing method with respect to appearance and geometry on a rendered video generated from the stylized scene. We had a sample of 31 users who anonymously perform the user study. 

\begin{table}[h]
\centering
\begin{tabular}{>{\centering\arraybackslash}p{1.6cm}  >{\centering\arraybackslash}p{0.6cm} >{\centering\arraybackslash}p{1.8cm} >{\centering\arraybackslash}p{0.8cm} >{\centering\arraybackslash}p{1.6cm}}
\toprule
Metric \quad ($\uparrow$) & \methodname (Ours) & Ref-NPR* \cite{zhang2023ref} & ARF* \cite{arf}  & S-DyRF \cite{li2024sdyrf} \\
\midrule
Preference & \textbf{41\%} & 5\% & 35\% & 19\% \\
\bottomrule
\end{tabular}
\caption{\textbf{User study results} where we compare the preference for the best performing method for a given scene and style image. We can observe that our method was preferred by most of the users, thereby verifying the quantitative and qualitative results.}
\label{tab:user_study}
\vspace{-0.25cm}
\end{table}


\section{Ablations}
We perform an extensive ablation study for supporting our design choices, especialy for the loss terms, gradient scaling, and the two step training process.

\begin{figure}[h]
     \centering
     \includegraphics[width=\linewidth]{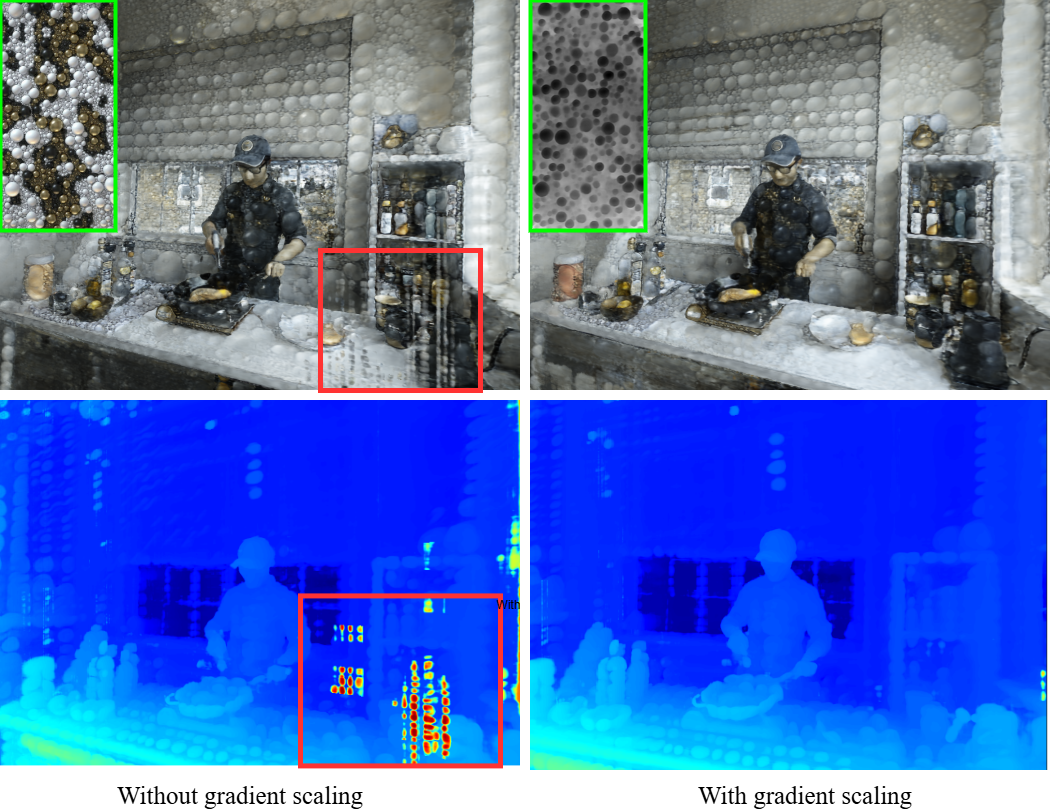}
     \caption{\textbf{Effect of gradient scaling} We demonstrate the effectiveness of the usage of linear gradient scaling, as explained in ~\Cref{sec:gradient_scaling_method}. Gradient scaling helps reduce the induced floater artifacts that arise while we are transferring the geometry from the style image onto our pretrained radiance field.}
     \label{fig:scalegrad}
 \end{figure}

\subsection{The effect of gradient scaling}
\label{sec:ablation:gradient_scaling}
To evaluate the impact of gradient scaling on our method, we conduct an ablation study by comparing results with and without gradient scaling. As shown in \Cref{fig:scalegrad}, the image on the left (without gradient scaling) exhibits noticeable artifacts, particularly in regions near the camera. In contrast, the image on the right (with gradient scaling) demonstrates a significant reduction in these artifacts. Gradient scaling helps stabilize the optimization process by preventing large gradient variations. This results in better visual quality, while also maintaining the required geometry.

\begin{figure}[h]
    \centering
    \includegraphics[width=\linewidth]{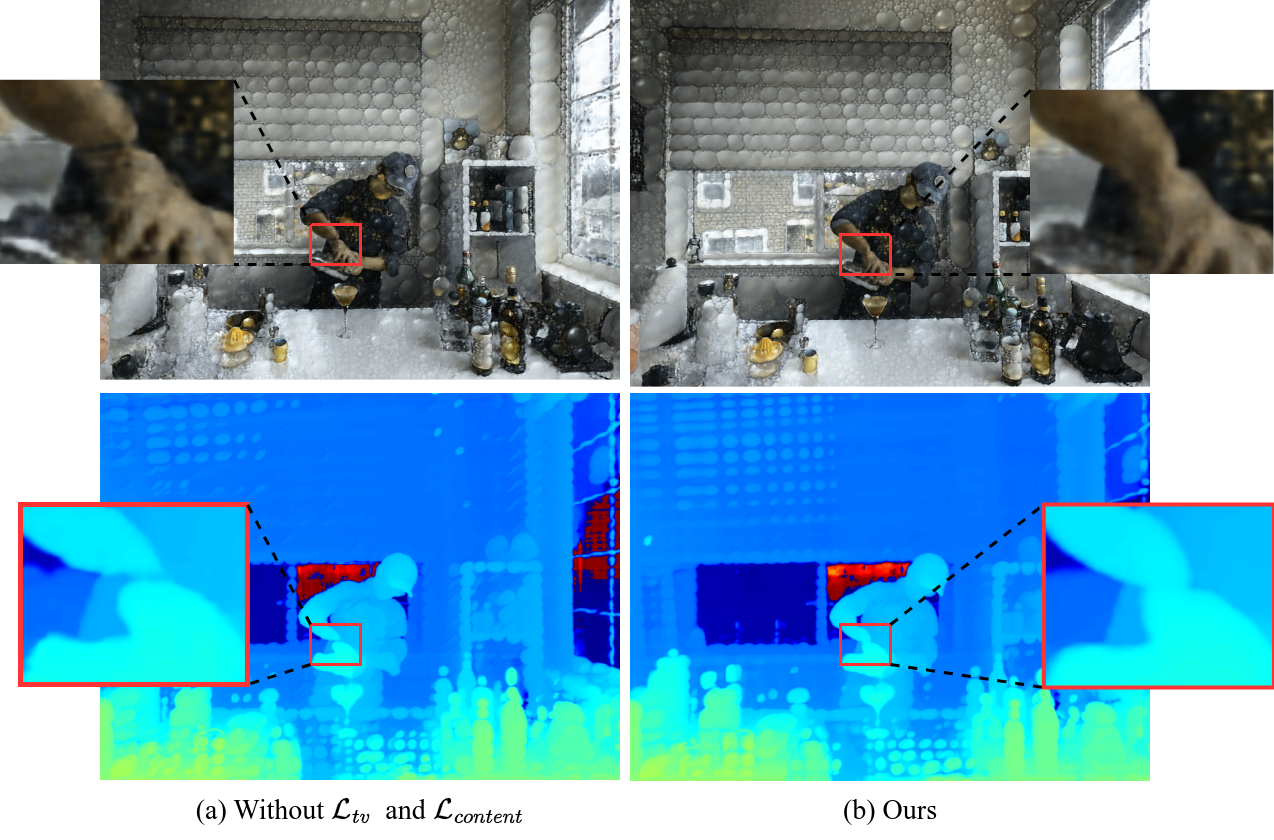}
    \caption{\textbf{Ablation of $\mathcal{L}_{tv}$ and $\mathcal{L}_{content}$}. We show the effect of the two loss functions. The purpose of these two loss functions was to maintain the presence of small details after the transfer of geometry to the stylized dynamic scene. It can be seen here that without the two loss terms, the hand of the human in the scene breaks into two parts, while on the right, the hand is able to be a single unit while also having the desired geometry effects.}
    \label{fig:broken_hand}
\end{figure}

\subsection{TV Loss and Depth Content loss}
\label{sec:ablation:tv_content_losses}
One basic idea, that is also explored for static scenes in ~\cite{3dgeo}, is to simply use the $NNFM$ loss for transferring the geometry of the style image onto the scene. Due to the formulation, it introduces a certain level of abstractness in the transfer of features from one feature map onto the other. While this is acceptable for color transfer, extra caution needs to be taken to maintain and preserve the geometrical details of the original underlying scene. Therefore, we introduced $\mathcal{L}_{tv}$ and $\mathcal{L}_{content}$ loss terms, 
When applying stylization without the $\mathcal{L}_{tv}$ and $\mathcal{L}_{content}$ losses, we observed a significant number of artifacts. However, by appropriately weighting the $\mathcal{L}_{tv}$ and $\mathcal{L}_{content}$, these artifacts became less prominent.
 
During geometry transfer, small regions may become overfitted to the style, leading to broken or distorted scenes. An example of this can be seen in \Cref{fig:broken_hand}(a), where the hand appears detached from the body. The incorporation of $\mathcal{L}_{tv}$ and $\mathcal{L}_{content}$ helps maintain the structural integrity of the scene, reducing such artifacts and preserving overall geometric consistency in the spatio-temporal domain.

\section{Conclusion}
We presented a novel method, \methodname, that achieves the task of applying style transfer, consisting of both appearance and geometry, to a dynamic radiance field. Our method follows a systematic approach to transfer the geometry, followed by the color transfer so as to have more coherence between the stylized appearance and the underlying geometry. We support the findings and demonstrate the effectiveness of \methodname through extensive experiments on synthetic and real-world data, and confirm the findings of manual choices by conducting an user study. We also support and inspect our architectural design choices by conducting ablation studies on core components and hence, their significance to the overall pipeline.

There are, however, many open challenges that could potentially be tackled in the future. Some of these include the use of a deep neural network to obtain test-time stylization without the need for computing an optimization for each given style image. Another possible direction is to work on a more strict coupling of appearance and geometry, with an increased focus on fitting the stylized appearance to a given geometry. To further facilitate research in this direction, we plan on releasing the code upon acceptance. 
{
    \small
    \bibliographystyle{ieeenat_fullname}
    \bibliography{main}

\begin{thebibliography}{62}
\providecommand{\natexlab}[1]{#1}
\providecommand{\url}[1]{\texttt{#1}}
\expandafter\ifx\csname urlstyle\endcsname\relax
  \providecommand{\doi}[1]{doi: #1}\else
  \providecommand{\doi}{doi: \begingroup \urlstyle{rm}\Url}\fi

\bibitem[Bhat et~al.(2023)Bhat, Birkl, Wofk, Wonka, and Müller]{zoedepth}
Shariq~Farooq Bhat, Reiner Birkl, Diana Wofk, Peter Wonka, and Matthias
  Müller.
\newblock Zoedepth: Zero-shot transfer by combining relative and metric depth,
  2023.

\bibitem[Cao and Johnson(2023)]{cao2023hexplane}
Ang Cao and Justin Johnson.
\newblock Hexplane: A fast representation for dynamic scenes.
\newblock In \emph{Proceedings of the IEEE/CVF Conference on Computer Vision
  and Pattern Recognition}, pages 130--141, 2023.

\bibitem[Chen et~al.(2017)Chen, Liao, Yuan, Yu, and Hua]{chen2017coherent}
Dongdong Chen, Jing Liao, Lu Yuan, Nenghai Yu, and Gang Hua.
\newblock Coherent online video style transfer.
\newblock In \emph{Proceedings of the IEEE International Conference on Computer
  Vision}, pages 1105--1114, 2017.

\bibitem[Chen et~al.(2022)Chen, Yuan, Li, Liu, Wang, Xie, Wen, and
  Yu]{chen2022upstnerf}
Yaosen Chen, Qi Yuan, Zhiqiang Li, Yuegen Liu, Wei Wang, Chaoping Xie, Xuming
  Wen, and Qien Yu.
\newblock Upst-nerf: Universal photorealistic style transfer of neural radiance
  fields for 3d scene, 2022.

\bibitem[Chiang et~al.(2022)Chiang, Tsai, Tseng, sheng Lai, and
  Chiu]{chiang2022stylizing}
Pei-Ze Chiang, Meng-Shiun Tsai, Hung-Yu Tseng, Wei sheng Lai, and Wei-Chen
  Chiu.
\newblock Stylizing 3d scene via implicit representation and hypernetwork,
  2022.

\bibitem[Chibane et~al.(2021)Chibane, Bansal, Lazova, and
  Pons-Moll]{chibane2021stereo}
Julian Chibane, Aayush Bansal, Verica Lazova, and Gerard Pons-Moll.
\newblock Stereo radiance fields (srf): Learning view synthesis for sparse
  views of novel scenes.
\newblock In \emph{Proceedings of the IEEE/CVF Conference on Computer Vision
  and Pattern Recognition}, pages 7911--7920, 2021.

\bibitem[Chou et~al.(2024)Chou, Huang, Liu, and
  Wang]{chou2024gsnerfgeneralizablesemanticneural}
Zi-Ting Chou, Sheng-Yu Huang, I-Jieh Liu, and Yu-Chiang~Frank Wang.
\newblock Gsnerf: Generalizable semantic neural radiance fields with enhanced
  3d scene understanding, 2024.

\bibitem[Deng et~al.(2022)Deng, Tang, Dong, Ma, Pan, Wang, and
  Xu]{deng2022stytr2imagestyletransfer}
Yingying Deng, Fan Tang, Weiming Dong, Chongyang Ma, Xingjia Pan, Lei Wang, and
  Changsheng Xu.
\newblock Stytr$^2$: Image style transfer with transformers, 2022.

\bibitem[Dosovitskiy et~al.(2015)Dosovitskiy, Fischer, Ilg, Hausser, Hazirbas,
  Golkov, Van Der~Smagt, Cremers, and Brox]{dosovitskiy2015flownet}
Alexey Dosovitskiy, Philipp Fischer, Eddy Ilg, Philip Hausser, Caner Hazirbas,
  Vladimir Golkov, Patrick Van Der~Smagt, Daniel Cremers, and Thomas Brox.
\newblock Flownet: Learning optical flow with convolutional networks.
\newblock In \emph{Proceedings of the IEEE international conference on computer
  vision}, pages 2758--2766, 2015.

\bibitem[Efros and Freeman(2023)]{efros2023image}
Alexei~A Efros and William~T Freeman.
\newblock Image quilting for texture synthesis and transfer.
\newblock In \emph{Seminal Graphics Papers: Pushing the Boundaries, Volume 2},
  pages 571--576, 2023.

\bibitem[Efros and Leung(1999)]{efros1999texture}
Alexei~A Efros and Thomas~K Leung.
\newblock Texture synthesis by non-parametric sampling.
\newblock In \emph{Proceedings of the seventh IEEE international conference on
  computer vision}, pages 1033--1038. IEEE, 1999.

\bibitem[Fridovich-Keil et~al.(2023)Fridovich-Keil, Meanti, Warburg, Recht, and
  Kanazawa]{kplanes_2023}
Sara Fridovich-Keil, Giacomo Meanti, Frederik~Rahbæk Warburg, Benjamin Recht,
  and Angjoo Kanazawa.
\newblock K-planes: Explicit radiance fields in space, time, and appearance.
\newblock In \emph{CVPR}, 2023.

\bibitem[Gao et~al.(2021)Gao, Saraf, Kopf, and Huang]{Gao2021DynNeRF}
Chen Gao, Ayush Saraf, Johannes Kopf, and Jia-Bin Huang.
\newblock Dynamic view synthesis from dynamic monocular video.
\newblock In \emph{Proceedings of the IEEE International Conference on Computer
  Vision}, 2021.

\bibitem[Garbin et~al.(2021)Garbin, Kowalski, Johnson, Shotton, and
  Valentin]{garbin2021fastnerf}
Stephan~J Garbin, Marek Kowalski, Matthew Johnson, Jamie Shotton, and Julien
  Valentin.
\newblock Fastnerf: High-fidelity neural rendering at 200fps.
\newblock \emph{arXiv preprint arXiv:2103.10380}, 2021.

\bibitem[Gatys et~al.(2016)Gatys, Ecker, and Bethge]{gatys2016image}
Leon~A Gatys, Alexander~S Ecker, and Matthias Bethge.
\newblock Image style transfer using convolutional neural networks.
\newblock In \emph{Proceedings of the IEEE conference on computer vision and
  pattern recognition}, pages 2414--2423, 2016.

\bibitem[Guangcong et~al.(2023)Guangcong, Chen, Loy, and
  Liu]{wang2023sparsenerf}
Guangcong, Zhaoxi Chen, Chen~Change Loy, and Ziwei Liu.
\newblock Sparsenerf: Distilling depth ranking for few-shot novel view
  synthesis.
\newblock \emph{IEEE/CVF International Conference on Computer Vision (ICCV)},
  2023.

\bibitem[Hedman et~al.(2021)Hedman, Srinivasan, Mildenhall, Barron, and
  Debevec]{hedman2021bakingneuralradiancefields}
Peter Hedman, Pratul~P. Srinivasan, Ben Mildenhall, Jonathan~T. Barron, and
  Paul Debevec.
\newblock Baking neural radiance fields for real-time view synthesis, 2021.

\bibitem[H{\"o}llein et~al.(2022)H{\"o}llein, Johnson, and
  Nie{\ss}ner]{hollein2022stylemesh}
Lukas H{\"o}llein, Justin Johnson, and Matthias Nie{\ss}ner.
\newblock Stylemesh: Style transfer for indoor 3d scene reconstructions.
\newblock In \emph{Proceedings of the IEEE/CVF Conference on Computer Vision
  and Pattern Recognition}, pages 6198--6208, 2022.

\bibitem[Huang et~al.(2017)Huang, Wang, Luo, Ma, Jiang, Zhu, Li, and
  Liu]{huang2017real}
Haozhi Huang, Hao Wang, Wenhan Luo, Lin Ma, Wenhao Jiang, Xiaolong Zhu, Zhifeng
  Li, and Wei Liu.
\newblock Real-time neural style transfer for videos.
\newblock In \emph{Proceedings of the IEEE conference on computer vision and
  pattern recognition}, pages 783--791, 2017.

\bibitem[Huang et~al.(2021)Huang, Tseng, Saini, Singh, and Yang]{huang2021lsnv}
Hsin-Ping Huang, Hung-Yu Tseng, Saurabh Saini, Maneesh Singh, and Ming-Hsuan
  Yang.
\newblock Learning to stylize novel views.
\newblock In \emph{Proceedings of the IEEE/CVF International Conference on
  Computer Vision}, 2021.

\bibitem[Huang and Belongie(2017)]{huang2017adain}
Xun Huang and Serge Belongie.
\newblock Arbitrary style transfer in real-time with adaptive instance
  normalization.
\newblock In \emph{Proceedings of the IEEE international conference on computer
  vision}, pages 1501--1510, 2017.

\bibitem[Huang et~al.(2022)Huang, He, Yuan, Lai, and
  Gao]{huang2022stylizednerf}
Yi-Hua Huang, Yue He, Yu-Jie Yuan, Yu-Kun Lai, and Lin Gao.
\newblock Stylizednerf: consistent 3d scene stylization as stylized nerf via
  2d-3d mutual learning.
\newblock In \emph{Proceedings of the IEEE/CVF Conference on Computer Vision
  and Pattern Recognition}, pages 18342--18352, 2022.

\bibitem[Ibrahimli et~al.(2024)Ibrahimli, Kooij, and
  Nan]{ibrahimli2024muviecast}
Nail Ibrahimli, Julian~FP Kooij, and Liangliang Nan.
\newblock Muviecast: Multi-view consistent artistic style transfer.
\newblock In \emph{2024 International Conference on 3D Vision (3DV)}, pages
  1136--1145. IEEE, 2024.

\bibitem[Jamri{\v{s}}ka et~al.(2019)Jamri{\v{s}}ka, Sochorov{\'a}, Texler,
  Luk{\'a}{\v{c}}, Fi{\v{s}}er, Lu, Shechtman, and
  S{\`y}kora]{jamrivska2019stylizingVideoByExample}
Ond{\v{r}}ej Jamri{\v{s}}ka, {\v{S}}{\'a}rka Sochorov{\'a}, Ond{\v{r}}ej
  Texler, Michal Luk{\'a}{\v{c}}, Jakub Fi{\v{s}}er, Jingwan Lu, Eli Shechtman,
  and Daniel S{\`y}kora.
\newblock Stylizing video by example.
\newblock \emph{ACM Transactions on Graphics (TOG)}, 38\penalty0 (4):\penalty0
  1--11, 2019.

\bibitem[Johnson et~al.(2016)Johnson, Alahi, and
  Fei-Fei]{johnson2016perceptual}
Justin Johnson, Alexandre Alahi, and Li Fei-Fei.
\newblock Perceptual losses for real-time style transfer and super-resolution.
\newblock In \emph{Computer Vision--ECCV 2016: 14th European Conference,
  Amsterdam, The Netherlands, October 11-14, 2016, Proceedings, Part II 14},
  pages 694--711. Springer, 2016.

\bibitem[Jung et~al.(2024)Jung, Nam, Sarafianos, Yoo, Sorkine-Hornung, and
  Ranjan]{3dgeo}
Hyunyoung Jung, Seonghyeon Nam, Nikolaos Sarafianos, Sungjoo Yoo, Alexander
  Sorkine-Hornung, and Rakesh Ranjan.
\newblock Geometry transfer for stylizing radiance fields.
\newblock In \emph{Proceedings of the IEEE/CVF Conference on Computer Vision
  and Pattern Recognition}, pages 8565--8575, 2024.

\bibitem[Kerbl et~al.(2023)Kerbl, Kopanas, Leimk{\"u}hler, and
  Drettakis]{kerbl3Dgaussians}
Bernhard Kerbl, Georgios Kopanas, Thomas Leimk{\"u}hler, and George Drettakis.
\newblock 3d gaussian splatting for real-time radiance field rendering.
\newblock \emph{ACM Transactions on Graphics}, 42\penalty0 (4), 2023.

\bibitem[Kov{\'a}cs et~al.(2024)Kov{\'a}cs, Hermosilla, and Raidou]{ggstyle}
{\'A}ron~Samuel Kov{\'a}cs, Pedro Hermosilla, and Renata~G Raidou.
\newblock G-style: Stylized gaussian splatting.
\newblock In \emph{Computer Graphics Forum}, page e15259. Wiley Online Library,
  2024.

\bibitem[Kwatra et~al.(2003)Kwatra, Sch{\"o}dl, Essa, Turk, and
  Bobick]{kwatra2003graphcut}
Vivek Kwatra, Arno Sch{\"o}dl, Irfan Essa, Greg Turk, and Aaron Bobick.
\newblock Graphcut textures: Image and video synthesis using graph cuts.
\newblock \emph{Acm transactions on graphics (tog)}, 22\penalty0 (3):\penalty0
  277--286, 2003.

\bibitem[Li et~al.(2022)Li, Slavcheva, Zollhoefer, Green, Lassner, Kim,
  Schmidt, Lovegrove, Goesele, Newcombe, et~al.]{li2022neural}
Tianye Li, Mira Slavcheva, Michael Zollhoefer, Simon Green, Christoph Lassner,
  Changil Kim, Tanner Schmidt, Steven Lovegrove, Michael Goesele, Richard
  Newcombe, et~al.
\newblock Neural 3d video synthesis from multi-view video.
\newblock In \emph{Proceedings of the IEEE/CVF conference on computer vision
  and pattern recognition}, pages 5521--5531, 2022.

\bibitem[Li et~al.(2024)Li, Cao, Wu, Wang, Xian, Wang, and Lin]{li2024sdyrf}
Xingyi Li, Zhiguo Cao, Yizheng Wu, Kewei Wang, Ke Xian, Zhe Wang, and Guosheng
  Lin.
\newblock S-dyrf: Reference-based stylized radiance fields for dynamic scenes.
\newblock In \emph{Proceedings of the IEEE/CVF Conference on Computer Vision
  and Pattern Recognition}, pages 20102--20112, 2024.

\bibitem[Liang et~al.(2024)Liang, Xu, Chen, Xiao, and
  Kang]{liang20244dstylegaussian}
Wanlin Liang, Hongbin Xu, Weitao Chen, Feng Xiao, and Wenxiong Kang.
\newblock 4dstylegaussian: Zero-shot 4d style transfer with gaussian splatting.
\newblock \emph{arXiv preprint arXiv:2410.10412}, 2024.

\bibitem[Liu et~al.(2023)Liu, Zhan, Chen, Zhang, Yu, El~Saddik, Lu, and
  Xing]{liu2023stylerf}
Kunhao Liu, Fangneng Zhan, Yiwen Chen, Jiahui Zhang, Yingchen Yu, Abdulmotaleb
  El~Saddik, Shijian Lu, and Eric~P Xing.
\newblock Stylerf: Zero-shot 3d style transfer of neural radiance fields.
\newblock In \emph{Proceedings of the IEEE/CVF Conference on Computer Vision
  and Pattern Recognition}, 2023.

\bibitem[Liu et~al.(2024)Liu, Zhan, Xu, Theobalt, Shao, and
  Lu]{liu2024stylegaussian}
Kunhao Liu, Fangneng Zhan, Muyu Xu, Christian Theobalt, Ling Shao, and Shijian
  Lu.
\newblock Stylegaussian: Instant 3d style transfer with gaussian splatting.
\newblock \emph{arXiv preprint arXiv:2403.07807}, 2024.

\bibitem[Liu et~al.(2021)Liu, Lin, He, Li, Wang, Li, Sun, Li, and
  Ding]{liu2021adaattnrevisitattentionmechanism}
Songhua Liu, Tianwei Lin, Dongliang He, Fu Li, Meiling Wang, Xin Li, Zhengxing
  Sun, Qian Li, and Errui Ding.
\newblock Adaattn: Revisit attention mechanism in arbitrary neural style
  transfer, 2021.

\bibitem[Luiten et~al.(2024)Luiten, Kopanas, Leibe, and
  Ramanan]{luiten2024dynamic}
Jonathon Luiten, Georgios Kopanas, Bastian Leibe, and Deva Ramanan.
\newblock Dynamic 3d gaussians: Tracking by persistent dynamic view synthesis.
\newblock In \emph{2024 International Conference on 3D Vision (3DV)}, pages
  800--809. IEEE, 2024.

\bibitem[Mei et~al.(2025)Mei, Xu, and Patel]{mei2025regs}
Yiqun Mei, Jiacong Xu, and Vishal Patel.
\newblock Regs: Reference-based controllable scene stylization with gaussian
  splatting.
\newblock \emph{Advances in Neural Information Processing Systems},
  37:\penalty0 4035--4061, 2025.

\bibitem[Mildenhall et~al.(2021)Mildenhall, Srinivasan, Tancik, Barron,
  Ramamoorthi, and Ng]{mildenhallnerf}
Ben Mildenhall, Pratul~P. Srinivasan, Matthew Tancik, Jonathan~T. Barron, Ravi
  Ramamoorthi, and Ren Ng.
\newblock Nerf: representing scenes as neural radiance fields for view
  synthesis.
\newblock \emph{Commun. ACM}, 65\penalty0 (1):\penalty0 99–106, 2021.

\bibitem[Mu et~al.(2022)Mu, Wang, Wu, and Li]{mu20223d}
Fangzhou Mu, Jian Wang, Yicheng Wu, and Yin Li.
\newblock 3d photo stylization: Learning to generate stylized novel views from
  a single image.
\newblock In \emph{Proceedings of the IEEE/CVF Conference on Computer Vision
  and Pattern Recognition}, pages 16273--16282, 2022.

\bibitem[M{\"u}ller et~al.(2022)M{\"u}ller, Evans, Schied, and
  Keller]{muller2022instant}
Thomas M{\"u}ller, Alex Evans, Christoph Schied, and Alexander Keller.
\newblock Instant neural graphics primitives with a multiresolution hash
  encoding.
\newblock \emph{ACM Transactions on Graphics (ToG)}, 41\penalty0 (4):\penalty0
  1--15, 2022.

\bibitem[Nguyen-Phuoc et~al.(2022)Nguyen-Phuoc, Liu, and
  Xiao]{nguyenphuoc2022snerf}
Thu Nguyen-Phuoc, Feng Liu, and Lei Xiao.
\newblock Snerf: Stylized neural implicit representations for 3d scenes.
\newblock \emph{ACM Trans. Graph.}, 2022.

\bibitem[Niemeyer et~al.(2021)Niemeyer, Barron, Mildenhall, Sajjadi, Geiger,
  and Radwan]{niemeyer2021regnerf}
Michael Niemeyer, Jonathan~T. Barron, Ben Mildenhall, Mehdi S.~M. Sajjadi,
  Andreas Geiger, and Noha Radwan.
\newblock Regnerf: Regularizing neural radiance fields for view synthesis from
  sparse inputs, 2021.

\bibitem[Park et~al.(2021{\natexlab{a}})Park, Sinha, Barron, Bouaziz, Goldman,
  Seitz, and Martin-Brualla]{park2021nerfies}
Keunhong Park, Utkarsh Sinha, Jonathan~T. Barron, Sofien Bouaziz, Dan~B
  Goldman, Steven~M. Seitz, and Ricardo Martin-Brualla.
\newblock Nerfies: Deformable neural radiance fields.
\newblock \emph{ICCV}, 2021{\natexlab{a}}.

\bibitem[Park et~al.(2021{\natexlab{b}})Park, Sinha, Hedman, Barron, Bouaziz,
  Goldman, Martin-Brualla, and Seitz]{park2021hypernerf}
Keunhong Park, Utkarsh Sinha, Peter Hedman, Jonathan~T. Barron, Sofien Bouaziz,
  Dan~B Goldman, Ricardo Martin-Brualla, and Steven~M. Seitz.
\newblock Hypernerf: A higher-dimensional representation for topologically
  varying neural radiance fields.
\newblock \emph{ACM Trans. Graph.}, 40\penalty0 (6), 2021{\natexlab{b}}.

\bibitem[Philip and Deschaintre(2023)]{nofloaters}
Julien Philip and Valentin Deschaintre.
\newblock Floaters no more: Radiance field gradient scaling for improved
  near-camera training.
\newblock \emph{arXiv preprint arXiv:2305.02756}, 2023.

\bibitem[Pumarola et~al.(2020)Pumarola, Corona, Pons-Moll, and
  Moreno-Noguer]{pumarola2020dnerf}
Albert Pumarola, Enric Corona, Gerard Pons-Moll, and Francesc Moreno-Noguer.
\newblock {D-NeRF: Neural Radiance Fields for Dynamic Scenes}.
\newblock In \emph{Proceedings of the IEEE/CVF Conference on Computer Vision
  and Pattern Recognition}, 2020.

\bibitem[Reiser et~al.(2021)Reiser, Peng, Liao, and Geiger]{reiser2021kilonerf}
Christian Reiser, Songyou Peng, Yiyi Liao, and Andreas Geiger.
\newblock Kilonerf: Speeding up neural radiance fields with thousands of tiny
  mlps, 2021.

\bibitem[Ruder et~al.(2018)Ruder, Dosovitskiy, and
  Brox]{ruder2018artisticStyleTransferVideos}
Manuel Ruder, Alexey Dosovitskiy, and Thomas Brox.
\newblock Artistic style transfer for videos and spherical images.
\newblock \emph{International Journal of Computer Vision}, 126\penalty0
  (11):\penalty0 1199--1219, 2018.

\bibitem[Saroha et~al.(2024)Saroha, Gladkova, Curreli, Muhle, Yenamandra, and
  Cremers]{saroha2024gaussian}
Abhishek Saroha, Mariia Gladkova, Cecilia Curreli, Dominik Muhle, Tarun
  Yenamandra, and Daniel Cremers.
\newblock Gaussian splatting in style.
\newblock \emph{arXiv preprint arXiv:2403.08498}, 2024.

\bibitem[Saroha et~al.(2025)Saroha, Hofherr, Gladkova, Curreli, Litany, and
  Cremers]{zdyss}
Abhishek Saroha, Florian Hofherr, Mariia Gladkova, Cecilia Curreli, Or Litany,
  and Daniel Cremers.
\newblock Zdyss--zero-shot dynamic scene stylization using gaussian splatting.
\newblock \emph{arXiv preprint arXiv:2501.03875}, 2025.

\bibitem[Simonyan and Zisserman(2014)]{simonyan2014vgg}
Karen Simonyan and Andrew Zisserman.
\newblock Very deep convolutional networks for large-scale image recognition.
\newblock \emph{arXiv preprint arXiv:1409.1556}, 2014.

\bibitem[Texler et~al.(2020)Texler, Futschik, Ku\v{c}era, Jamri\v{s}ka,
  \v{S}\'{a}rka Sochorov\'{a}, Chai, Tulyakov, and S\'{y}kora]{texlerfspbt}
Ond\v{r}ej Texler, David Futschik, Michal Ku\v{c}era, Ond\v{r}ej Jamri\v{s}ka,
  \v{S}\'{a}rka Sochorov\'{a}, Menglei Chai, Sergey Tulyakov, and Daniel
  S\'{y}kora.
\newblock Interactive video stylization using few-shot patch-based training.
\newblock \emph{ACM Transactions on Graphics}, 39\penalty0 (4):\penalty0 73,
  2020.

\bibitem[Wang et~al.(2023)Wang, Liu, Chen, Liu, Liu, Komura, Theobalt, and
  Wang]{wang2023f2nerffastneuralradiance}
Peng Wang, Yuan Liu, Zhaoxi Chen, Lingjie Liu, Ziwei Liu, Taku Komura,
  Christian Theobalt, and Wenping Wang.
\newblock F$^{2}$-nerf: Fast neural radiance field training with free camera
  trajectories, 2023.

\bibitem[Wei and Levoy(2000)]{wei2000fast}
Li-Yi Wei and Marc Levoy.
\newblock Fast texture synthesis using tree-structured vector quantization.
\newblock In \emph{Proceedings of the 27th annual conference on Computer
  graphics and interactive techniques}, pages 479--488, 2000.

\bibitem[Wu et~al.(2024)Wu, Yi, Fang, Xie, Zhang, Wei, Liu, Tian, and
  Wang]{wu20244dgs}
Guanjun Wu, Taoran Yi, Jiemin Fang, Lingxi Xie, Xiaopeng Zhang, Wei Wei, Wenyu
  Liu, Qi Tian, and Xinggang Wang.
\newblock 4d gaussian splatting for real-time dynamic scene rendering.
\newblock In \emph{Proceedings of the IEEE/CVF Conference on Computer Vision
  and Pattern Recognition}, pages 20310--20320, 2024.

\bibitem[Xu et~al.(2024)Xu, Chen, Xiao, Sun, and Kang]{xu2024styledyrf}
Hongbin Xu, Weitao Chen, Feng Xiao, Baigui Sun, and Wenxiong Kang.
\newblock Styledyrf: Zero-shot 4d style transfer for dynamic neural radiance
  fields.
\newblock \emph{arXiv preprint arXiv:2403.08310}, 2024.

\bibitem[Yang et~al.(2024)Yang, Gao, Zhou, Jiao, Zhang, and
  Jin]{yang2024deformable}
Ziyi Yang, Xinyu Gao, Wen Zhou, Shaohui Jiao, Yuqing Zhang, and Xiaogang Jin.
\newblock Deformable 3d gaussians for high-fidelity monocular dynamic scene
  reconstruction.
\newblock In \emph{Proceedings of the IEEE/CVF Conference on Computer Vision
  and Pattern Recognition}, pages 20331--20341, 2024.

\bibitem[Yu et~al.(2021{\natexlab{a}})Yu, Li, Tancik, Li, Ng, and
  Kanazawa]{yu2021plenoctreesrealtimerenderingneural}
Alex Yu, Ruilong Li, Matthew Tancik, Hao Li, Ren Ng, and Angjoo Kanazawa.
\newblock Plenoctrees for real-time rendering of neural radiance fields,
  2021{\natexlab{a}}.

\bibitem[Yu et~al.(2021{\natexlab{b}})Yu, Ye, Tancik, and
  Kanazawa]{yu2021pixelnerfneuralradiancefields}
Alex Yu, Vickie Ye, Matthew Tancik, and Angjoo Kanazawa.
\newblock pixelnerf: Neural radiance fields from one or few images,
  2021{\natexlab{b}}.

\bibitem[Zhang et~al.(2022)Zhang, Kolkin, Bi, Luan, Xu, Shechtman, and
  Snavely]{arf}
Kai Zhang, Nick Kolkin, Sai Bi, Fujun Luan, Zexiang Xu, Eli Shechtman, and Noah
  Snavely.
\newblock Arf: Artistic radiance fields.
\newblock In \emph{European Conference on Computer Vision}, pages 717--733.
  Springer, 2022.

\bibitem[Zhang et~al.(2018)Zhang, Isola, Efros, Shechtman, and
  Wang]{zhang2018unreasonable}
Richard Zhang, Phillip Isola, Alexei~A Efros, Eli Shechtman, and Oliver Wang.
\newblock The unreasonable effectiveness of deep features as a perceptual
  metric.
\newblock In \emph{Proceedings of the IEEE conference on computer vision and
  pattern recognition}, pages 586--595, 2018.

\bibitem[Zhang et~al.(2023)Zhang, He, Xing, Yao, and Jia]{zhang2023ref}
Yuechen Zhang, Zexin He, Jinbo Xing, Xufeng Yao, and Jiaya Jia.
\newblock Ref-npr: Reference-based non-photorealistic radiance fields for
  controllable scene stylization.
\newblock In \emph{Proceedings of the IEEE/CVF Conference on Computer Vision
  and Pattern Recognition}, pages 4242--4251, 2023.

\end{thebibliography}
}

\end{document}